\documentclass{article}

\usepackage[english]{babel}

\usepackage[letterpaper,top=2cm,bottom=2cm,left=3cm,right=3cm,marginparwidth=1.75cm]{geometry}

\usepackage{amsmath}
\usepackage{csquotes}
\usepackage{graphicx}
\usepackage[colorlinks=true, allcolors=blue]{hyperref}

\title{Reasonable Scale Machine Learning with Open-Source Metaflow}
\author{Jacopo Tagliabue (New York University), Hugo Bowne-Anderson (Outerbounds),  \\ Ville Tuulos (Outerbounds), Savin Goyal (Outerbounds),  \\Romain Cledat (Netflix), David Berg (Netflix)\footnote{\textit{Contributions}: Ville, Savin, Romain, David, and the extended team at Netflix are co-creators of Metaflow; Jacopo and Hugo led the writing process for this article; Jacopo co-created Metaflow cards with the Outerbounds team. Everybody was involved in writing the final draft. Please note that Jacopo was at \textit{Coveo} when the paper was first written.}}

\begin{document}
\maketitle

\begin{abstract}

As Machine Learning (ML) gains adoption across industries and new use cases, practitioners increasingly realize the challenges around effectively developing and iterating on ML systems: reproducibility, debugging, scalability, and documentation are elusive goals for real-world pipelines outside tech-first companies. In \textit{this} paper, we review the nature of ML-oriented workloads and argue that re-purposing existing tools won't solve the current productivity issues, as ML peculiarities warrant specialized development tooling. We then introduce \texttt{Metaflow}, an open-source framework for ML projects explicitly designed to boost the productivity of data practitioners by abstracting away the execution of ML code from the definition of the business logic. We show how our design addresses the main challenges in ML operations (MLOps), and document through examples, interviews and use cases its practical impact on the field. 

\end{abstract}
\section{Introduction}
\label{sec:intro}

\begin{quote}
``We shape our tools and then our tools shape us'' -- (probably not) M. McLuhan.
\end{quote}

Machine learning (ML) models are increasingly important in our digital (and physical) life, through recommenders \cite{45530,https://doi.org/10.48550/arxiv.2209.05310}, search engines \cite{Mohan2019,bianchi-etal-2021-query2prod2vec}, digital helpers \cite{10.1145/3292500.3330723}, as well as countless other applications. Due to strategic advantages in data wealth, computational capabilities, and culture, the first adopters of ML have been tech-first companies \cite{NIPS2012_6aca9700,Bennett2007TheNP}. However, in the coming years, the majority of ML deployment will be motivated by all sorts of use cases, either internal to companies \cite{gartnerenterprise}, or consumer-facing at \textit{a non-planetary scale} \cite{10.1145/3475965.3479315}: Machine Learning ``at Reasonable Scale'' is now a popular phrase \cite{towardDataScienceJT,ericMLOPSMess} to cover ML development along the mid and long tail of the market, as opposed to the bespoke requirements in both tooling and infrastructure necessary at Big Tech scale. \textit{Reasonable Scale} itself denotes quite a big range and, as we shall see below, applies to use cases more than organizations: from data science workflows to training large research-grade ML models, we have seen Reasonable Scale teams handle datasets from a few thousands rows to tens of GB of data.

Unfortunately, the ML workflow at Reasonable Scale appears to often be broken: when surveyed, practitioners admit projects take months to go into production, and more than half the time they actually never make it past the experimental stage \cite{gartnerproduction}. In parallel, the data-centric AI movement has been increasingly vocal on the importance of the operational context in which model operates \cite{ReGithub,NG}: as the flexibility of deep learning points towards commoditization of model building \cite{DBLP:journals/corr/abs-1909-07930}, the remaining parts in the ML workflow remain challenging for the average practitioner \cite{10.1145/3411764.3445518,43146}: reproducibility, debugging, scalability, and even documentation are elusive goals \cite{Bailis2017InfrastructureFU}. As an example, recent surveys report that 96\% of enterprises still encounter challenges in managing the life-cycle of data-based workflows \cite{datasurvery}.

At the organizational level, the Reasonable Scale is characterized by agile teams, speed over-optimization, and re-use of available tooling (as opposed to bespoke infrastructure) \cite{10.1145/3460231.3474604}. This has direct implications at the use case level: borrowing the three categories for data systems from \cite{stonebraker12}, we can see how success does not hinge on \textit{large} petabyte scale problems, nor on the \textit{highest velocity} deployments with minimal latency. Instead, success comes from the flexibility of solving the largest variety of use cases through an improved development experience. In \textit{this} paper, we motivate the reasons and design choices behind \texttt{Metaflow}, an open-source framework explicitly designed for ML projects. We summarize our contributions as follows:

\begin{enumerate}
    \item we provide an overview of a typical ML workflow and best practices and motivate the need for an ML-specific tool as opposed to re-purposing generic libraries or cloud platforms;
    \item we present \texttt{Metaflow} key ideas in a principled way and show how its design answers common challenges for practitioners;
    \item we document successful \texttt{Metaflow} deployments from early adopters, showcase the range of applications the framework already supports, and present a case study in team topology to highlight the organizational benefits of proper ML tooling.
\end{enumerate}

Developing in the open for both industry and academic practitioners puts us in a privileged position to observe the fast-evolving field of MLOps: in our final section, we sketch our vision for the future of \texttt{Metaflow} and present our reflections on the changing landscape of ML engineering as a profession.

\section{Background and motivation}
\label{sec:background}
The rate of adoption of ML in the last few years has been tremendous: according to a recent survey, 75\% of CTOs have ML as a top priority \cite{algoritmia2021}. If ML initiatives abound, successes have been however highly skewed, with few companies reaping the most of ML benefits: \cite{algo2019} reported that for half of their enterprise sample, it takes more than a month to deploy a model into production, while only 14\% could do so in less than a week; \cite{hbr2022} reports that 20\% or fewer models reach production deployment. 

Increasing demand for data scientists\footnote{There is considerable debate in the field about the exact responsibilities of \textit{data scientists}, \textit{machine learning engineers}, \textit{machine learning scientists}, and so on. As it will become increasingly clear, \texttt{Metaflow} supports natively \textit{any} Python workload based on \textit{data and code}: for this reason, we will use all the above terms interchangeably in \textit{this} paper.} opened the gate to the software industry to many professionals \cite{hbr2012}, whose main skill is not software as a craft, but instead data exploration and statistical modeling\footnote{For example, the US Bureau of Labor Statistics estimates for data scientists that the percent change in employment from 2021 to 2031 is 21\%, versus a national average of 5\%.}. Companies at the Reasonable Scale need therefore to optimize productivity with two conflicting constraints: teams need to be small, but the available data talent may lack engineering and infrastructure knowledge, making it hard to own features end to end. As a result, the ``handoff'' became a popular team topology: the team that builds the model is not the one building the production pipeline. While no topology is right or wrong \textit{per se}, there is a significant consensus among data and software leaders that such a pattern does not set them up for success: as recently highlighted for example in \cite{skelton2019team}, it is crucial that teams ``own the software'' end-to-end, and avoid that ``multiple teams change the same system''. The importance of good tools cannot, therefore, be overestimated, as more investments are poured into ML initiatives: by making it easy to do the right thing -- through good abstractions and a pleasant coding experience -- tools can help small teams become productive irrespective of their infrastructure abilities (see the case study in Section \ref{sec_case}).

A second important theme for ML productivity has been the rise of cloud computing -- cheaper, accessible, ubiquitous -- and the exponential improvement of local workstations. Notwithstanding the historical importance of the ``Big Data'' era and tooling \cite{5496972,10.5555/2228298.2228301}, many use cases in current ML do \textit{not} require horizontal scaling: after initial data wrangling is performed on data warehouses, the resulting training sets are often small enough to fit into memory; moreover, memory on modern laptops and off-the-shelf cloud boxes has dramatically increased, making vertical scaling particularly effective across a wide variety of scenarios, including data-intensive application such as e-commerce \cite{bianchi2020bert}. While there is of course ample room for special, case-by-case considerations, it is telling that almost all the paradigmatic ``Big Data'' datasets from \cite{doi:10.1089/big.2013.0037} -- or even recent Data Challenges tailored on deep learning approaches \cite{CoveoSIGIR2021,https://doi.org/10.48550/arxiv.2207.05772} -- can be comfortably wrangled locally \cite{10.1145/3299869.3320212} or in a single cloud instance\footnote{At the moment of writing this paper, a single X1 machine on AWS comes with 2 TB of memory and 128 CPUs.}. As forcefully argued by the seminal COST paper \cite{McSherry2015ScalabilityBA}, good non-distributed implementations are often very competitive across a wide range of scenarios. In sub-fields like NLP, swapping between CPU and GPU tasks is often more central to practitioners' day-to-day tasks than swapping between one-node and multi-node clusters. 

\subsection{Existing libraries and tools}
As recognized by most practitioners, the logical unit of ML work is the \textit{pipeline}, and not the \textit{model}. Even if, by magic, modeling was reduced to a single line of code, a project going from data to serving would still comprise many other steps \cite{43146,DBLP:journals/corr/abs-2110-13601}. 

As ML pipelines are a combination of code and data, many companies, especially those adopting ML as market pioneers, re-purpose existing orchestrators to serve as the backbone of their ML pipelines: given that Python - with major libraries such as scikit-learn \cite{sklearn_api}, PyTorch \cite{NEURIPS2019_9015}, Tensorflow \cite{tensorflow2015-whitepaper} - is the language of choice for most practitioners, picking Python-based orchestrators (e.g. Luigi \cite{luigi}, Airflow \cite{airflow} and Prefect \cite{prefect}) is a convenient heuristics when choosing between a multitude of options\footnote{Orchestrators come in many shapes and sizes: see, for example, \url{https://github.com/common-workflow-language/common-workflow-language/wiki/Existing-Workflow-systems}.}. While re-purposing a Python orchestrator is low-hanging fruit, the ergonomics of these libraries tend to be low for ML-first workload: Luigi, for example, does not support data artifacts as first-class citizens, and Airflow users often experience a wide and costly gap between experimentation and production (Section \ref{sec:interviews}).

Another popular strategy has been developing on top of the Kubernetes ecosystem \cite{43826}, such as Kubeflow by Google\footnote{\url{https://www.kubeflow.org/}}, Flyte by Lyft\footnote{\url{https://flyte.org/}}, and Argo Workflows by Intuit\footnote{\url{https://argoproj.github.io/argo-workflows/}}. While these systems have many desirable characteristics from the engineering point of view -- they are robust, scalable, and highly-available -- they inherit the complexity endemic in Kubernetes, making them less usable by data scientists with limited infrastructure experience. On the other hand, decoupling logic from the actual engine executing the code is a fundamental design principle for \texttt{Metaflow}, which aims to abstract away infrastructural details from the developer experience of data practitioners (Section \ref{sec:ideas}).

Finally, all major cloud providers have been very active in the space, with AzureML\footnote{\url{https://azure.microsoft.com/en-us/products/machine-learning/}}, Sagemaker\footnote{\url{https://aws.amazon.com/it/pm/sagemaker}} and Vertex AI\footnote{\url{https://cloud.google.com/vertex-ai}. We can also note that Vertex came to market significantly later than \texttt{Metaflow}: 2021 vs 2019.} providing training and serving capabilities through UI or cloud SDK. While convenient, full-cloud based solutions are not ideal either for many practitioners who wish to have control over their workloads: they are fairly opinionated in data exchanges, with high vendor lock-in, they are not open source and typically cloud-only, making for longer feedback loops and slower iterations compared to local development.

\subsection{Why a new tool?}
There are two main philosophical reasons to develop a new open source tool for ML pipelines: one is the core belief that ``working in ML'' is fundamentally different from data pipelines and software engineering; the other is the opportunity in the market of ideas for an offering which truly caters to \textit{most} practitioners in the middle of the tail - i.e. those who don't want a no-code solution but still don't have the custom needs of Big Tech.

\begin{figure*}
  \centering
  \includegraphics[width=\textwidth]{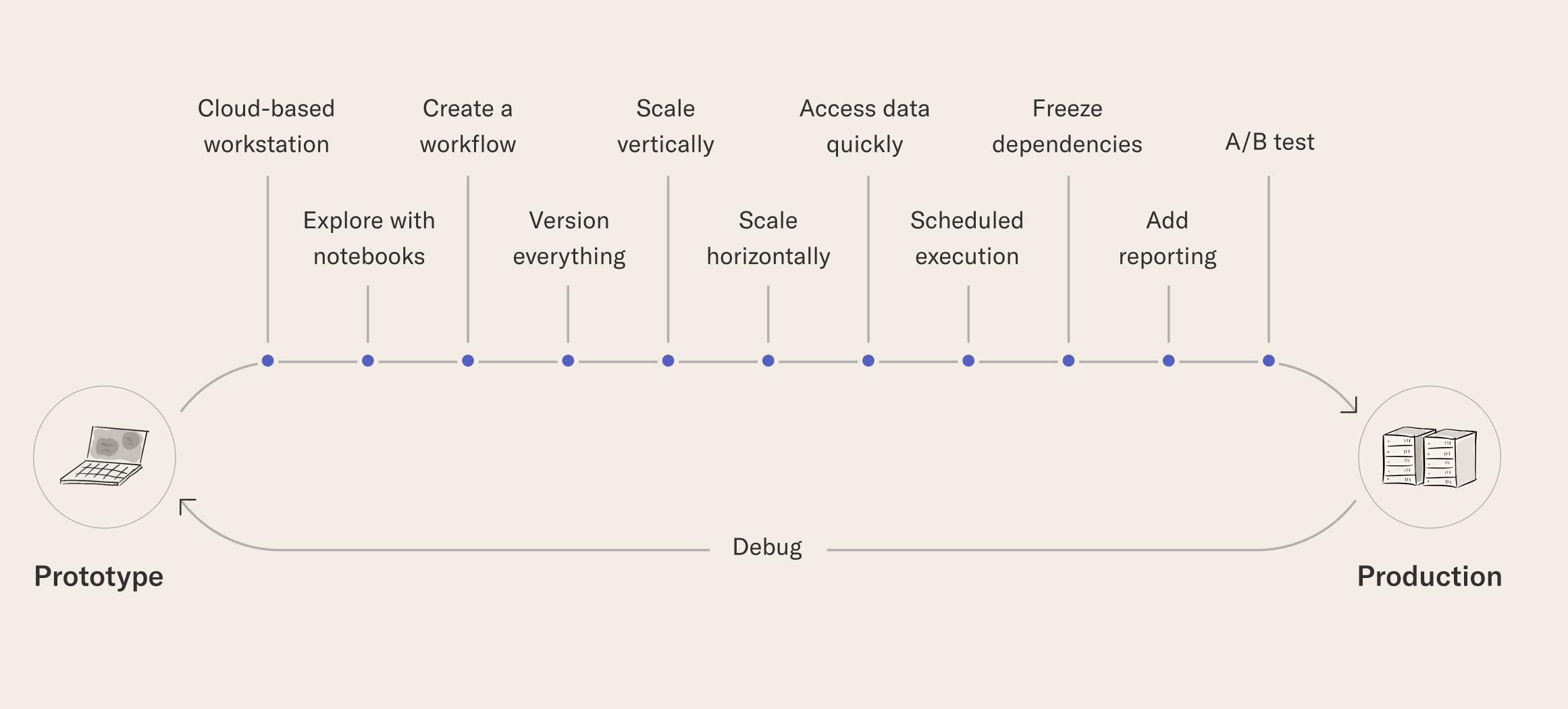}
  \caption{\textbf{An ML workflow, from prototype to production}. Since data science challenges are often open-ended, running several experiments before deciding on the final configuration is not uncommon. Streamlining the workflow to access production-level resources (such as data and compute), safely versioned, and in the same cloud environment will result in significant productivity gains. Note also the arrow pointing \textit{backwards}: debugging and iteration should ideally happen within the same operational context (a similar workflow is also depicted in \cite{https://doi.org/10.48550/arxiv.2209.09125}).}
  \label{fig:workflow}
  \vspace{-4mm}
\end{figure*}

The peculiarity of the data science workflow is illustrated in Fig. \ref{fig:workflow}. The line from Prototype to Production is rarely just a line, as it's not uncommon to go back and forth between different data samples and modeling techniques before settling on a final configuration \cite{VilleBook}. The question is \textit{not} whether it is possible, through incremental changes and constant refactoring, to make custom Python code scale, run on a schedule, and interact with datasets: it certainly is, using a combination of the tools and techniques, and with the support of different teams (Section \ref{sec:interviews}). The real challenge is to empower data scientists to move between all these tasks while \textit{never} changing code and fully within a declarative paradigm.

As some have argued, the open-ended nature of data work and the rapid iterations required for success make the adoption of standard software best practices sometimes an obstacle to productivity \cite{https://doi.org/10.48550/arxiv.2209.09125}. Shipping incrementally is still a best practice, but ``incrementally'' doesn't just mean ``a small code change at a time'', but may include variations in sampling, pre-processing, training scheduled, and/or modeling libraries. Whatever the right dosage of software best practices is, fast iterations and streamlined deployments are keys to success: a data scientist should be able to move from local coding to scheduled inference, to cloud debugging in a seamless way. The boundaries between ``local and cloud'' blur indefinitely, as \textit{production-ready is indeed a continuum} \cite{villeyoutube}: exploration is never ``done'' on one side, but real-world feedback ``can’t come soon enough'' on the other. 

\texttt{Metaflow} has been designed from the ground up to be a tool for ML productivity: common concerns are abstracted away, infrastructure is declarative, and moving from local to cloud runs (and back) is seamless (see Section \ref{sec:ideas}). It is important to note that \texttt{Metaflow} was born at \textit{Netflix} to cater to use cases \textit{inside the company}: even Big Tech companies with sophisticated users can still choose standardized abstractions when the use case does not depend on a bespoke infrastructure. In other words, even companies with planetary scale ML workloads \textit{also} have many reasonable scale pipelines to deal with, often for internal stakeholders, suggesting that the ``Reasonable Scale'' label properly applies to use cases more than to organizations.

After increasing adoption in many different types of companies (see Section \ref{sec:examples}), \texttt{Metaflow} has been battle tested in prototype, research, and production workloads across a wide variety of ML scenarios. 

In a recent survey, \cite{https://doi.org/10.48550/arxiv.2209.09125} summarized in \textit{velocity}, \textit{validation}, \textit{versioning} the main properties that ``dictate how successful deployments will be'': their own main takeaway is that ``there's an opportunity to create virtualized infrastructure specific to ML needs with similar development and production environments. Each environment should build on the same foundation but supports different modes of iteration (i.e., high velocity in development).''. We shall see in the ensuing sections how \texttt{Metaflow} addresses these \textit{desiderata}.

\section{Metaflow: Main ideas}
\label{sec:ideas}

ML pipelines are modeled as Directed Acyclic Graphs (DAG), i.e. graphs where each node represents a step (e.g. getting data, training a model) and a directed edge $E_{ij}$ links two steps, $i$ and $j$, if $j$ depends for its execution on the successful termination of $i$. As a real-world example, we could consider the pipeline from \cite{Chia2021AreYS} (also depicted in Fig. 1 in \cite{Tagliabue2021DAGCI}), in which a DAG transforms raw e-commerce data (images and html tables) into a cached representation of product substitutes (e.g. a LED TV will have other similar TVs as substitutes). 

There are five main concepts that come together to build such a DAG:

\begin{enumerate}
    \item \textbf{Dependencies}: what depends on what, e.g. model training depends on first successfully transforming the dataset; note how some tasks don't need to wait on others: image and html processing can and should happen in parallel;
    \item \textbf{Business Logic}: the code specifying the task logic, e.g. the code training the Siamese network; 
    \item \textbf{Environment}: the (Python) environment running the code, e.g. a \textit{mini-conda} environment, a \textit{virtualenv}, or a fully specified Docker container;
    \item \textbf{Computation}: where and when the code is run (triggered and on a schedule), e.g. a local workstation or a GPU-enabled machine in the cloud;
    \item \textbf{Documentation}: where we communicate what the pipeline does, e.g. DAG cards \cite{Tagliabue2021DAGCI}.
\end{enumerate}

\begin{figure*}
  \centering
  \includegraphics[width=\textwidth]{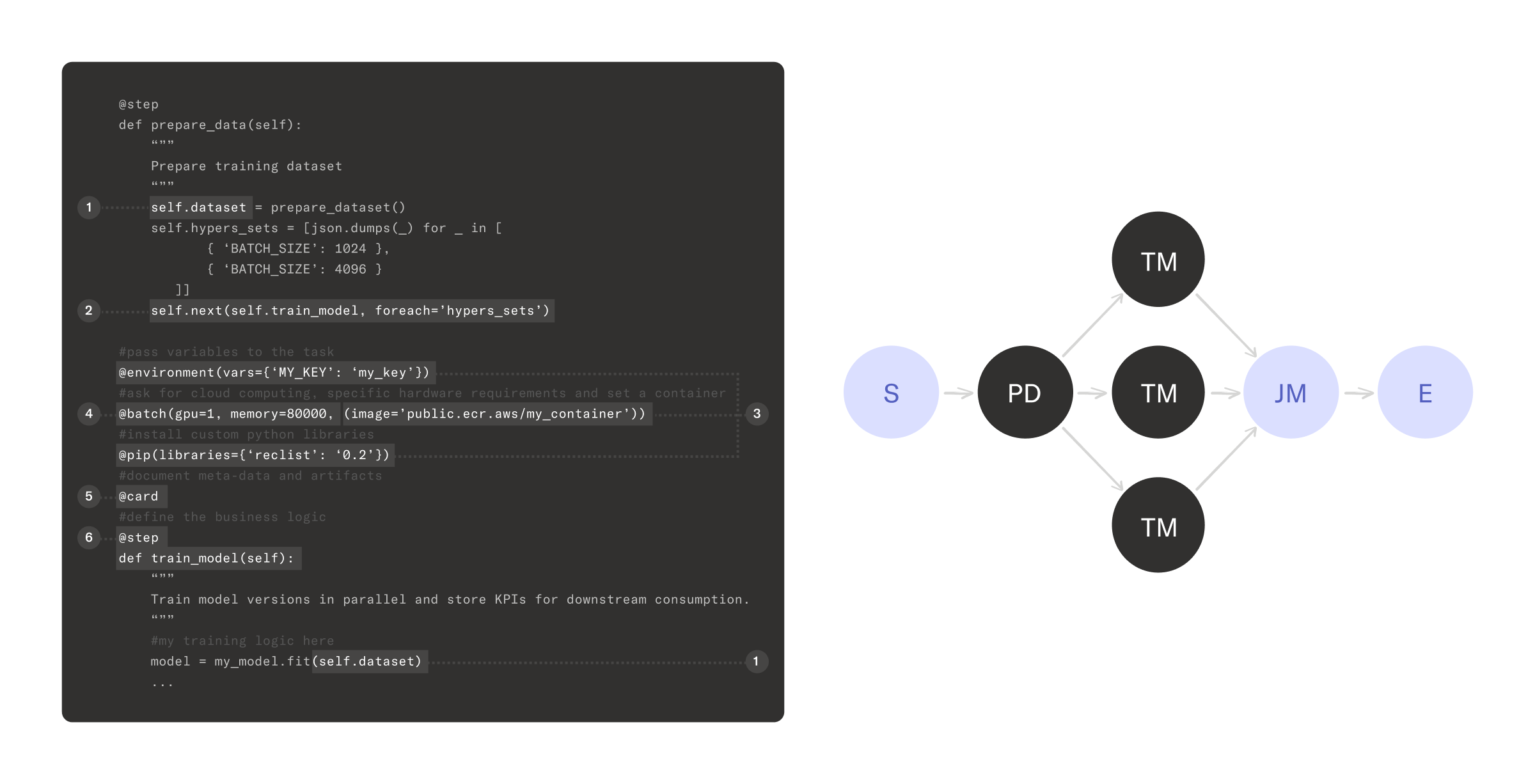}
  \caption{\textbf{Code snippet}. A stripped-down DAG and its representation with \texttt{Metaflow}: \textbf{S}tart, \textbf{P}repare \textbf{D}ataset, \textbf{T}rain \textbf{M}odel (in parallel), \textbf{J}oin \textbf{M}odel, \textbf{E}nd (black circles are the steps visible in the Python snippet). 
  \texttt{Metaflow} allows the definition of all main ML workflow concepts through idiomatic Python, including infrastructure and documentation. \textit{Legend}. \textit{1}: storing and loading datasets; \textit{2} defining task dependencies and parallelism; \textit{3} defining the Python environment, packages and variables; \textit{4} defining computing location and hardware requirements; \textit{5} self-documenting meta-data and artifacts when running.}
  \label{fig_code}
  \vspace{-4mm}
\end{figure*}

Fig. \ref{fig_code} shows a snippet from \texttt{Metaflow} illustrating the above concepts for a toy ML pipeline (\textit{right}). Before going into the details, note that all \textit{concepts live together in code}, defined independently (i.e. they can all be changed without affecting other settings). No special files or configurations are needed, as the library leverages the expressivity of Python and the declarative nature of decorators to provide the desired functionalities:

\begin{enumerate}
    \item \textbf{Dependencies} (\textbf{2}): a special command \texttt{next} defines the dependency between $prepare\_data$ and $train\_model$. Note the \texttt{foreach} argument, which allows the parallel execution of $n$ training tasks, one with each of the hyperparameter settings specified in the list passed as value;
    \item \textbf{Business Logic} (\textbf{6}): logic is defined through standard Python functions, decorated with the \texttt{step} decorator; since we are allowed arbitrary Python code, the model sophistication is up to the practitioner, from simple parametric models to fine-tuning of large multi-modal neural networks \cite{Naturearticle};
    \item \textbf{Environment} (\textbf{3}): the characteristics of the runtime -- environmental variables, containerized dependencies, selective pip installs -- are all controlled through code;
    \item \textbf{Computation} (\textbf{4}): the \texttt{batch} decorator packages, ships, and runs through AWS Batch (or Kubernetes) only the relevant task code. Through arguments, it is possible to specify hardware requirements that the runner needs to satisfy; in this example, instead of running an entire pipeline on an expensive GPU only because one specific task requires it, the data scientist can run locally or on inexpensive boxes all the pre-processing tasks and just use on-demand compute for the training step without any code change or explicit infrastructure work;
    \item \textbf{Documentation} (\textbf{5}): the \texttt{card} decorator will generate a shareable artifact documenting the pipeline structure and metadata, and out-of-the-box visualization for some variables.
\end{enumerate}

The developer experience provided by \texttt{Metaflow} is the consequence of three design choices:

\begin{enumerate}
    \item \textit{everything-as-code}: logical dependencies, Python packages, and Docker images are all declared in Python;
    \item \textit{declarative infrastructure}: parallelization of compute, provisioning of specialized hardware (i.e. GPUs), and support for cloud workloads is purely declarative: users specify \textit{what} needs to happen, \texttt{Metaflow} transparently figures out \textit{how};
    \item \textit{native versioning}: models and datasets are automatically snapshotted and versioned through the idiomatic \texttt{self} syntax (\textit{1}, in Fig. \ref{fig_code}).
\end{enumerate}

Taken together, these choices are \texttt{Metaflow}'s response to the three key challenges of successful ML deployments \cite{https://doi.org/10.48550/arxiv.2209.09125}:

\begin{itemize}
    \item \textbf{Velocity}: \texttt{Metaflow} improve iterations speed on several axes: it removes DevOps dependencies from data science teams\footnote{Note that one-line \textit{scheduling} for DAGs as a workflow with workflow orchestrators like Airflow, Argo Workflows, AWS Step Functions is another important feature of \texttt{Metaflow}, not shown in Fig. \ref{fig_code}.}; it decouples logic from computation, allowing local execution for quick feedback and cloud execution when needed.
    \item \textbf{Validation}: \texttt{Metaflow} helps with building confidence in pipelines by providing many out-of-the-box functionalities: linting and code checks, breakpoints for interactive debugging, and, importantly the possibility of seamless debugging cloud DAGs \textit{locally} through the centralized data store (see below) -- it is easy to reproduce the work of other team members or develop parallel versions of the production DAGs. Importantly, \texttt{Metaflow}'s Python-first approach favors interoperability with popular testing tools for data and ML, such as \textit{Great Expectations} and \textit{RecList} \cite{gong_abe_2022_7094191,10.1145/3487553.3524215}.
    \item \textbf{Versioning}: every variable in a DAG run can be automatically versioned, namespaced by users, in a data store (e.g. model weights, datasets, parameters). Everybody with the proper access can retrieve the exact state of the system from \textit{any} past run, in development or production environments, and reproduce errors locally. Through the \texttt{@card} decorator (\textit{5} in Fig.\ref{fig_code}), \texttt{Metaflow} also produces a DAG card as versioned documentation of a given run. 
\end{itemize}

Having described \texttt{Metaflow}'s key principles, and how they are meant to address the peculiarity of ML development, the ensuing section explains how the implementation choices shaped the final developer experience.

\section{Implementation details}
\label{sec:details}

\texttt{Metaflow} is available as an open source Python package released under the Apache License 2.0\footnote{While we have mostly directed our attention to Python-based ecosystems, it is worth noting that \texttt{Metaflow} also comes with bindings for R.}. By design, decoupling logic and dependencies from computation and storage makes \texttt{Metaflow} compatible with a wide range of ``back-end'' solutions: a fully local deployment, a cloud-backed deployment (through AWS, Azure, and Google Cloud services), and a custom, Docker-based deployment (through Kubernetes) -- as mentioned, given the abstractions, developers can move from one to the other seamlessly, significantly lowering migration cost and vendor lock-in.

We provide implementation details at two different levels of abstractions: first, we assume the developer experience perspective, and discuss the semantics of \texttt{Metaflow} main concepts; second, we assume the system design perspective and explain how runtime and orchestration work behind the scenes to provide such a developer experience.

\subsection{Lexicon}

\texttt{Metaflow} distinguishes between the abstract DAG representing the ML pipeline and the actual execution(s) of the pipeline. Developers use the \texttt{@step} decorator to mark functions as \textbf{Steps} in the pipeline, and organize them in a \textbf{Flow}, the structure of dependencies between functions that make up the DAG (Fig. \ref{fig_code}). When the code is executed, a \textbf{Run} is the execution of the flow, made up from \textbf{Tasks}, which are the executions of a step. The product of running flows is data, models, and other \emph{immutable} data \textbf{Artifacts}: as mentioned, every run is versioned as unique, and so are its task and artifacts. In other words, if the code is run twice, \texttt{Metaflow} will internally distinguish between the two runs and give granular access to artifacts if necessary.

\subsection{Architecture}

\begin{figure}
  \centering
  \includegraphics[width=8cm]{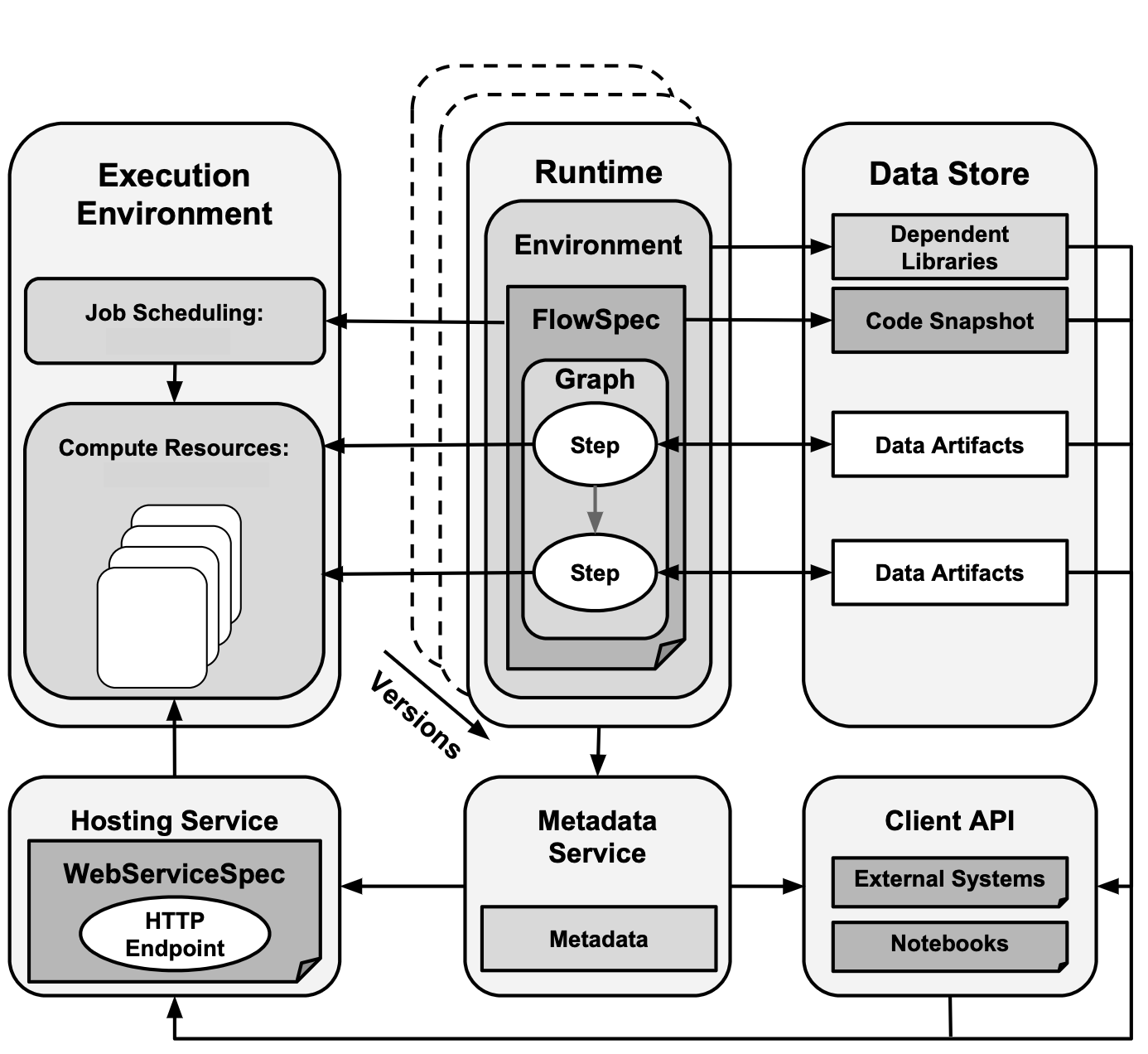} 
  \caption{\texttt{Metaflow} architecture.}
  \label{fig_architecture}
\end{figure}

Fig. \ref{fig_architecture} provides a high-level architectural summary of its primary components, which we now survey.

\textit{Runtime}. The runtime is responsible for orchestrating and coordinating the reproducible execution of a workflow. The runtime schedules step execution on either the local instance or the cloud by acquiring compute resources from the execution environment, gathering and storing user code, data artifacts, and runtime environment in the data store, and tracking and logging information about the execution to the metadata service. By abstracting away the computational layer from the business logic, \texttt{Metaflow} blurs the distinction between local and cloud development, avoiding costly context-switch between environments. Additionally, Metaflow makes passing state downwards from one step to the other trivial by ensuring that instance variables are logged in a persisted store and made lazily available later.

\textit{Execution environment}. The execution environment is responsible for provisioning compute resources for individual steps of the workflow. For rapid prototyping, \texttt{Metaflow} provides a local execution engine that runs each step in a separate process (not shown). Remote execution of steps, whether initiated manually or scheduled via a scheduler, takes place through the cloud provider, completely transparently for the developer. At the moment of writing \textit{this} paper, \texttt{Metaflow} supports Amazon Web Services, Microsoft Azure, and Google Cloud Platform as cloud providers and Kubernetes-based stack for custom solutions.

\textit{Data Store}. \texttt{Metaflow} requires an object store to persist user code, data artifacts, and runtime environment. Cloud storage services (e.g. S3) provide a highly durable and available data store that is accessible from all execution environments where code executes. ML pipelines may need to read and write large dataframes as variables that get re-used throughout the flow: using storage services affords us the ability to decouple compute from storage, and potentially make available artifacts to various consumers even outside of \texttt{Metaflow} (e.g. Presto, Spark, DuckDB). To prevent data-duplicity, \texttt{Metaflow} utilizes content-addressable storage as its storage mechanism. Since data access is so important, \texttt{Metaflow} provides a custom implementation for S3 access, optimized for high throughput. The library is capable of saturating the network connection on the largest instances that provide throughput of up to 25Gbps. Achieving this level of throughput needs careful management of multiple TCP connections over multiple processes, which can't be straightforwardly achieved using the standard Python library for AWS Boto\footnote{Available at \url{https://boto3.amazonaws.com/v1/documentation/api/latest/index.html}.}.

\textit{Client}. \texttt{Metaflow} client (also in Python) allows for accessing user code, data artifacts, runtime environment, and execution statistics of past executions of a flow within a notebook, another workflow, or any other user code. Since runs are automatically tagged with user names, an important side-effect of deterministic versioning is that the client lets users access artifacts and results from other people running experiments with the same flow: runs are uniquely tagged and isolated at runtime, to avoid interfering with other people's pipelines, but they are also fully accessible afterward in a uniform way through the client, allowing team members to compare insights and share artifacts. 

\textit{Metadata Service}. A centralized metadata service keeps track of workflow executions.

As a result of its design, \texttt{Metaflow} allows users to deterministically transfer the state (code, variables, dependencies) of any run between environments, to ensure reproducibility and streamlined debugging. A typical use case is detecting an anomaly in the last run of a \textit{production} flow: a \texttt{Metaflow} developer could retry the flow locally (which is now automatically tagged as a \textit{user} run) and debug it \textit{as if} it were in production, but without actually conflicting with the existing pipeline or asking for the intervention of other teams. As mentioned in Section \ref{sec:background}, one of the main motivations behind \texttt{Metaflow} is making small, focused data science teams productive end-to-end.

\subsection{Running a flow: a worked-out example}

In the following, we outline the steps that take place when a user executes a flow locally on their workstation:

\begin{enumerate}
\item \emph{Validate the user-defined flow}. This ensures that the Python class represents a well-formed DAG and notifies the user immediately if issues are detected.
\item \emph{Package code for execution}. To make it possible to execute
steps in remote environments, e.g. on AWS Batch or Kubernetes, all Python files are packaged and stored in the data store.
\item \emph{Begin graph traversal from the start step}. Every well-formed \texttt{Metaflow} DAG has an unambiguous \texttt{start} step which starts the run.
\item \emph{Launch a supervisor process for a remote task}. If a task is marked for remote execution, a local supervisor process is started to handle the interaction with the cloud.
\item \emph{Execute a task}. A separate OS-level process is started to
execute the user-defined code in a step as an individual task. The task process can either execute locally or in a remote environment set up in step 4.
\item \emph{Persist artifacts}. After the user code completes, \texttt{Metaflow} inspects \texttt{self} attributes created by the user, serializes them, and persists them in the content-addressed datastore unless an existing copy of the artifact exists already.
\item \emph{Proceed to the next step}. Thanks to step 1, we know that we are executing a valid DAG and hence we can introspect the next step, annotated with \texttt{self.next}. The algorithm returns to step 4, unless the current step is the \texttt{end} step which ends the execution.
\end{enumerate}

Notice how this algorithm interleaves elements of \emph{workflow orchestration} i.e. graph traversal (steps 1, 3, 7), as well as \emph{computation} (steps 2, 4, 5) and \emph{data flow} (step 6).

A crucial feature of \texttt{Metaflow} is that internally orchestration and compute can be decoupled, although the user needs to interact only with one coherent API. Besides interactive local execution described above, flows can be deployed in production where they run automatically without any human interaction, with orchestration by a highly available, scalable orchestrator like AWS Step Functions, Argo Workflows, or Airflow.

In the production case, steps (1, 3, 7) are managed by the external orchestrator and the rest of the steps are managed by \texttt{Metaflow}. This ensures that the user's \emph{business logic}, \emph{environment}, and \emph{documentation} stay intact when moving from prototype to production, while a highly available orchestrator ensures that the flow is always executed reliably on time.

\section{Adoption and deployment scenarios}
\label{sec:examples}

As a testament to its versatility, \texttt{Metaflow} is used by practitioners in large traditional corporations (e.g. \textit{CNN}), Big Tech (\textit{Netflix}), AI startups (\textit{Shaped}, \textit{Kailua Labs}) and industrial research groups (\textit{Coveo AI}).

While acknowledging how hard it is to prove the value of productivity tooling as compared to, say, benchmarking ML models, we organize evidence from different sources to highlight the real-world impact of \texttt{Metaflow}: the role of the library in the general open source ecosystem; pain points and interviews with practitioners; a case study with a \textit{before} vs \textit{after} comparison, through the lens of an ML leader at the target company \footnote{Please note that the practitioners have affiliations and titles as of June 2022, when we conducted the initial user study: they may not necessarily reflect the current roles.}.
 
\subsection{Open source community}

As of this moment, \texttt{Metaflow} has over 6000 GitHub stars. Known use cases solved with \texttt{Metaflow} span a variety of industries and modeling techniques:

\begin{itemize}
    \item \textit{23andMe}, a company working on consumer-scale genetic kits, provides reports on conditions like Type 2 Diabetes and HDL Cholesterol \cite{23andme}.
    \item \textit{Netflix}, a company in the entertainment industry, determines the launch calendar for new movies, solving a combinatorial optimization problem with a set of constraints.
    \item \textit{CNN}, a media company, designs personalized experiences for users searching for fresh news. Since moving their data science to \texttt{Metaflow}, they were able to ``test twice as many models in Q1 2021 as they did in all of 2020'' \cite{cnn}.
    \item The research group at \textit{Coveo}, an API company providing ML models as a service, built replicable and scalable pipelines for research-grade experimentation, resulting in papers at top-tier ML venues \cite{Chia2021AreYS,Naturearticle}. 
    \item \textit{Realtor.com}, a website featuring popular home search destinations, builds adaptive marketing campaigns by embedding data scientists in the marketing team \cite{realtor}.
\end{itemize}

\subsection{Practitioners interviews}
\label{sec:interviews}

We interviewed three practitioners with hands-on leadership roles in ML, who are using \texttt{Metaflow} for Reasonable Scale use cases fitting a wide range of organization complexity: a startup, a small tech public company (less than 1BN USD in market cap), and a large public tech company (over 100B USD in market cap). We asked each of them to describe their previous pipeline stack, pain points, and how moving to \texttt{Metaflow} addressed them.

\subsubsection{Startup: Shaped}

\textit{Tullie Murrell} is the CEO of \textit{Shaped}, a startup providing cutting-edge RecSys models as a service (through APIs) to its clients. \textbf{Original stack}: Airflow as the pipeline orchestrator for \textit{production}, AWS Sagemaker (from the command line) for experimental workflows, AWS Sagemaker as compute engine. \textbf{Pain points}: scalability issues with Airflow when multiple copies of the same pipeline need to be scheduled independently; coupling of logic and computation to use Sagemaker effectively; wide gap between experimentation flow and production, as well as between local and remote runs; sharing data between Airflow operators is slow and not intuitive. \textbf{New stack}: \texttt{Metaflow} reduces the gap between experiments and production, as the same code now runs on different compute engines on demand; it prevents vendor lock-in through decoupling of logic and execution; it streamlines data sharing between steps, environments and runs through the S3 Data Store abstraction.

\subsubsection{Small Public Company: Coveo}

\textit{Ciro Greco} is the Vice-President of AI at Coveo, a small public company, which provides ML models in the Information Retrieval space to mid and large enterprises, with a SaaS business model. \textbf{Original stack}: Airflow as the pipeline orchestrator for \textit{production}, Luigi as the glue for the ML pipeline, AWS EC2 and Fargate as compute engine. \textbf{Pain points}: provisioning GPUs was manual and expensive, as the entire Luigi pipeline now runs on the GPU; moving data between steps with Luigi is possible, but only locally with respect to the script: no deterministic versioning exists; dependency and debugging issues when running the pipeline locally and then remotely as part of another tool (Airflow). \textbf{New stack}: \texttt{Metaflow} allows to selectively launch GPUs for training, while leaving the rest of the pipeline on commodity hardware, either locally or remotely; centralized data versioning and data sharing through the client API simplify data management, and streamlines connection with other AWS tool (e.g. a model serialized in S3 can be immediately deployed on Sagemaker endpoints using purely Python code\footnote{For a working example, see for example the deep learning RecSys pipeline at \url{https://github.com/jacopotagliabue/recs-at-resonable-scale}.}); team ramp-up is reduced to a single tool, and there is no dependency on data and DevOps team to launch and debug production-level flows.

\subsubsection{Large Public Company: Netflix}

The Netflix ML platform supports a wide range of Python-based data science and machine learning workflows, with applications across the company from recommendations to demand modeling and media. \textbf{Original stack}: A wide variety of internal and OSS company-managed tools for low-level access to cloud development, compute, scheduling, dependencies, notebooks, and data access. \textbf{Pain points}: Before \texttt{Metaflow}, data scientists were spending a lot of time managing the complexity of engineering tools and differences in environments. They would spend weeks in slow development cycles and have more difficulties when going into production because the tools and documentation at their disposal were designed to make platform engineers productive, not data scientists. \textbf{New stack}: Users no longer have to manage the complexity of the different environments and tools and have increased operational independence and productivity. Data scientists are now able to fully productionize their pipelines in hours to days instead of weeks. The Netflix ML platform is working to integrate with more production systems to provide a single, consistent user experience through \texttt{Metaflow}'s plugins. Thanks to its open source nature, The Netflix ML platform was also able to extend the functionality of \texttt{Metaflow}, and make it more cohesive with their company's environment, for example, implementing model hosting and fast data access patterns.

\subsection{Case study}
\label{sec_case}

\begin{figure}
  \centering
  \includegraphics[width=8cm]{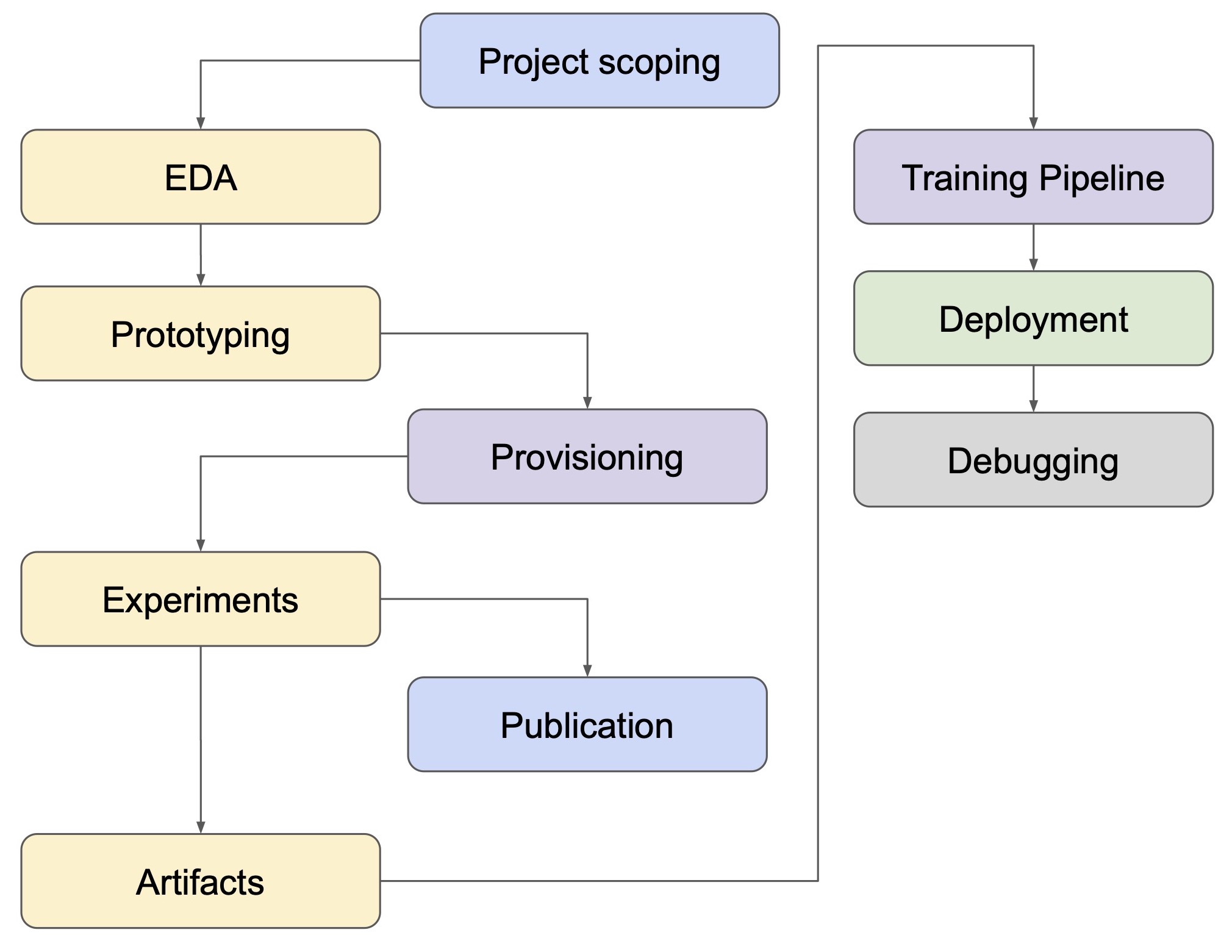} 
  \caption{Workflow \textit{before} the adoption of \texttt{Metaflow}, from the use case in Section \ref{sec_case}. \textit{Blue} tasks are for the PI, \textit{yellow} tasks for the data scientist, \textit{purple} tasks for the data engineering team, \textit{green} tasks for the API team, \textit{gray} tasks shared among all actors.}
  \label{fig_before}
\end{figure}

\textit{Jacopo Tagliabue} is the founder and Principle Investigator (PI) of \textit{Coveo AI}, an applied R\&D lab, providing both product innovation and research-quality publications to a public company \cite{jacopo_medium}. His stated goal is to improve the state of the art of some important use cases for the company (session-based inference \cite{tagliabue-etal-2021-bert}, multi-modal recommendations \cite{chia-etal-2022-come} etc.) \textit{and} then productionize the model as part of the usual API offering. As Section \ref{sec:interviews} was focused on tools and pain points,  here we take the organizational perspective, to illustrate how \texttt{Metaflow} can change team topology through better ownership of the development process.

Fig. \ref{fig_before} depicts a typical functional workflow for an ML project (e.g. improving session-based recommendations for a certain use case) before the introduction of \texttt{Metaflow}: different colors highlight different personas involved at different stages. In particular:

\begin{itemize}
    \item data scientists (yellow) need to wait for data engineers (purple) for the provisioning of appropriate resources for full-scale training after prototyping has been completed (e.g. GPU machines); however, data engineers need to wait for data scientists to coalesce their local pipeline of scripts in artifacts, in order to port it to the production job scheduler;
    \item data engineers are now responsible to produce fresh model artifacts for deployment, introducing a new dependency in the flow: the API team (green), responsible for the customer-facing endpoint, is now dependent on the data engineers for up-to-date models, but data engineers \textit{are not the ones that built the model in the first place};
    \item when something goes wrong, debugging (grey) is a shared responsibility among all the actors: as reproducing issues is tantamount to walking the flow backward, team dependencies slow the process down and can lead to unhealthy team dynamics. 
\end{itemize}

\begin{figure}
  \centering
  \includegraphics[width=6cm]{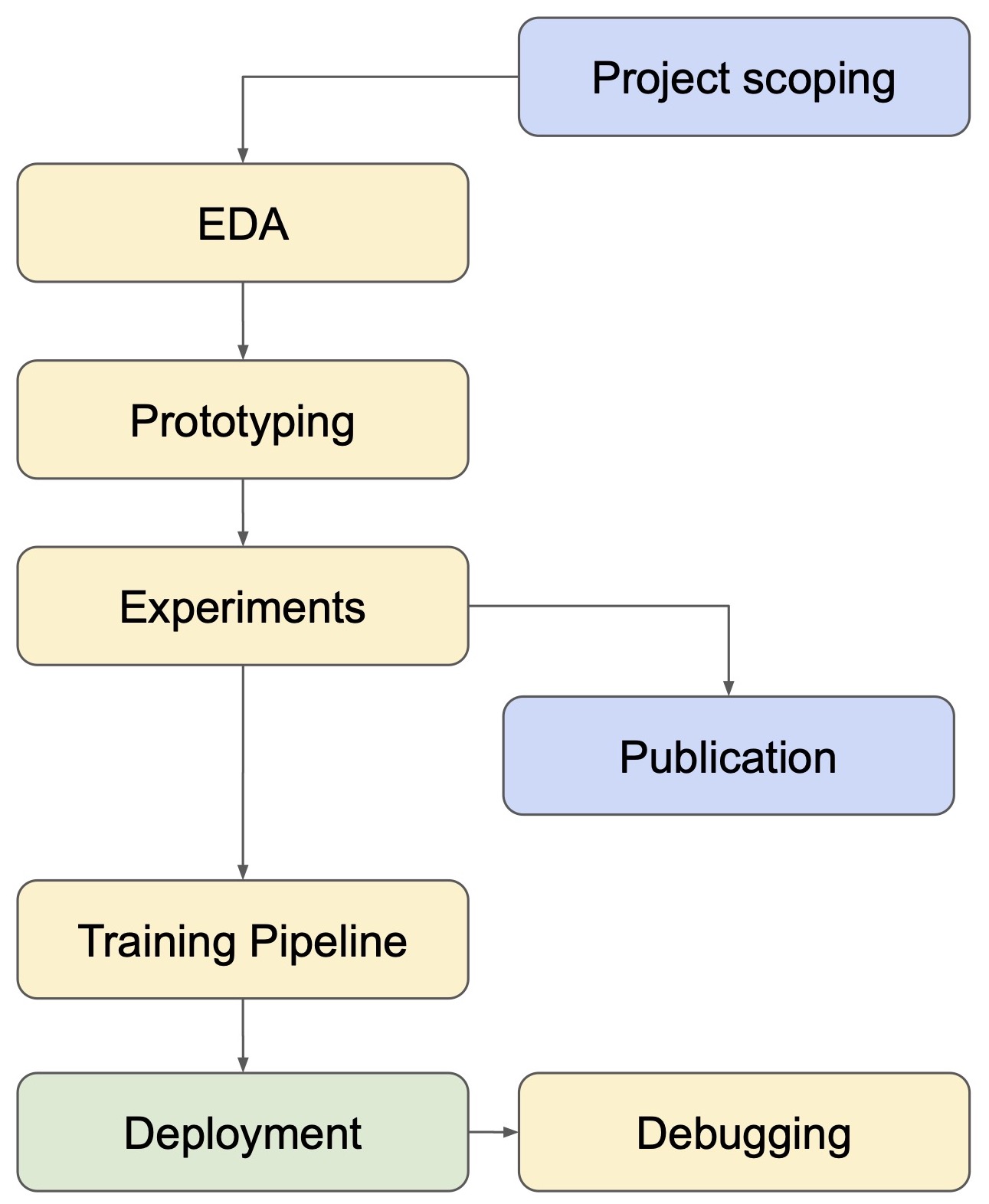} 
  \caption{Workflow simplification \textit{after} the adoption of \texttt{Metaflow}, Section \ref{sec_case}.}
  \label{fig_after}
\end{figure}

Fig. \ref{fig_after} is the same workflow, after the introduction of \texttt{Metaflow} to abstract away infrastructure and versioning for the data science team (Section \ref{sec:ideas}). \textit{Provisioning} is now unnecessary, thanks to the declarative nature of infrastructure provisioning; equally important, built-in scheduling with no code changes removes the need for data engineers in the training pipeline. Since artifacts are produced automatically through versioning, the API team has now a healthy, direct dependency on the data science team: the same pipeline is owned at the experiment, training, and debugging phase by one team, which is empowered to make logic and library decisions independently from the other teams. A shared cloud-based bucket for artifacts, such as S3, is the only implicit contract between the data science team and the rest of the engineering organization.

Based on our experience, the organizational constraints of \textit{Coveo AI} are representative of a large class of companies in which machine learning development is both necessary \textit{and hard}. While some problems are (and will remain) ``people problems'' more than ``tooling problems'', by focusing on the productivity of data scientists, we believe \texttt{Metaflow} abstractions can also have significant benefits at the organizational level, promoting a culture of ownership and fast iteration \cite{dataOpsblog}. 

\section{Conclusion and future work}

In \textit{this} paper, we presented \texttt{Metaflow}, an open source tool designed to enhance the developer experience and productivity of data science teams. We motivated the rationale behind the tool, and showed how its abstractions and implementation are a direct result of a deep understanding of practitioners' problems, as data scientist dissatisfaction is increasingly recognized in the field (see \cite{https://doi.org/10.48550/arxiv.2209.09125,ericMLOPSMess}, and also our own interviews in Section \ref{sec:interviews}).

As the future of ML deployments will take place, largely, at Reasonable Scale, lowering the bar for production-ready pipelines will also help alleviate the current talent shortage \cite{10.1145/3460231.3474604}. As modeling gets mostly commoditized \cite{10.1145/3475965.3479315}, ML-first tooling like \texttt{Metaflow} will play a larger role in democratizing access to Machine Learning and spread its benefits even outside of large tech companies.

We anticipate that while the models and the sizes of datasets won't grow exponentially in Reasonable Scale organizations, the overall level of sophistication will keep increasing. We want to stay responsive to the changing needs of practitioners. Along these lines, our future work includes flow composition, i.e., support for increasingly sophisticated ML systems which consist of not only one, but many interconnected workflows, as well as more advanced configuration management which such systems require. We want to support also larger GPU and CPU clusters, as well as lower the latency of interactive executions to speed up development cycles even further.

On top of this, as the authors of \cite{https://doi.org/10.48550/arxiv.2209.09125} astutely point out, ``We see this as opportunities for new resources, such as classroom material (e.g., textbooks, courses) to prescribe the right engineering practices and rigor for the highly experimental discipline that is production ML.'' To this end we are dedicated to writing books, developing up-to-date technical resources for practitioners, and building development environments, such as our \texttt{Metaflow} sandbox, that lower the barrier to entry for everybody to take part in MLOps.

\subsection{How to get started}

For the reader's convenience, we report links to code, tutorials, and additional resources to get started with \texttt{Metaflow}:

\begin{itemize}
    \item \texttt{Metaflow} is available as a fully open source project (Apache License 2.0) at: \url{https://github.com/Netflix/metaflow};
    \item a free sandbox allows to try \texttt{Metaflow} through a web-based interface and transparent cloud-based deployment: \url{https://outerbounds.com/docs/sandbox/};
    \item \cite{VilleBook} is book-length treatment of \texttt{Metaflow} in the context of data science workflows, supported by open source code exercises and examples: \url{https://github.com/outerbounds/dsbook};
    \item a growing list of tutorials for different ML use cases (NLP, computer vision, RecSys, etc.) is available at: \url{https://outerbounds.com/docs/tutorials-index/}. 
\end{itemize}

\section*{Acknowledgments}

\texttt{Metaflow} was originally developed at \textit{Netflix}, and it is now further developed by \textit{Netflix}, Outerbounds, and other companies. We thank the practitioners that generously gave their time for our user study and four anonymous reviewers for feedback on a previous version of the paper. Finally, thanks to \texttt{Metaflow} users: some we know about, many we don't, but they all motivate us to make the framework better every day.

\bibliographystyle{alpha}
\bibliography{sample}

\newcommand{\etalchar}[1]{$^{#1}$}
\begin{thebibliography}{TTGD21b}

\bibitem[AAB{\etalchar{+}}15]{tensorflow2015-whitepaper}
Mart\'{i}n Abadi, Ashish Agarwal, Paul Barham, Eugene Brevdo, Zhifeng Chen,
  Craig Citro, Greg~S. Corrado, Andy Davis, Jeffrey Dean, Matthieu Devin,
  Sanjay Ghemawat, Ian Goodfellow, Andrew Harp, Geoffrey Irving, Michael Isard,
  Yangqing Jia, Rafal Jozefowicz, Lukasz Kaiser, Manjunath Kudlur, Josh
  Levenberg, Dandelion Man\'{e}, Rajat Monga, Sherry Moore, Derek Murray, Chris
  Olah, Mike Schuster, Jonathon Shlens, Benoit Steiner, Ilya Sutskever, Kunal
  Talwar, Paul Tucker, Vincent Vanhoucke, Vijay Vasudevan, Fernanda Vi\'{e}gas,
  Oriol Vinyals, Pete Warden, Martin Wattenberg, Martin Wicke, Yuan Yu, and
  Xiaoqiang Zheng.
\newblock {TensorFlow}: Large-scale machine learning on heterogeneous systems,
  2015.
\newblock Software available from tensorflow.org.

\bibitem[AGH{\etalchar{+}}22]{https://doi.org/10.48550/arxiv.2209.05310}
Rohan Anil, Sandra Gadanho, Da~Huang, Nijith Jacob, Zhuoshu Li, Dong Lin, Todd
  Phillips, Cristina Pop, Kevin Regan, Gil~I. Shamir, Rakesh Shivanna, and Qiqi
  Yan.
\newblock On the factory floor: Ml engineering for industrial-scale ads
  recommendation models, 2022.

\bibitem[{Alg}21]{algoritmia2021}
{Algorithmia Research}.
\newblock 2021 enterprise trends in machine learning, 2021.

\bibitem[{Apa}22]{airflow}
{Apache Foundation}.
\newblock Apache airflow, 2022.

\bibitem[Asa17]{gartnerproduction}
Matt Asay.
\newblock 85\% of big data projects fail, but your developers can help yours
  succeed, 2017.

\bibitem[BL07]{Bennett2007TheNP}
James Bennett and Stan Lanning.
\newblock The netflix prize.
\newblock 2007.

\bibitem[BLB{\etalchar{+}}13]{sklearn_api}
Lars Buitinck, Gilles Louppe, Mathieu Blondel, Fabian Pedregosa, Andreas
  Mueller, Olivier Grisel, Vlad Niculae, Peter Prettenhofer, Alexandre
  Gramfort, Jaques Grobler, Robert Layton, Jake VanderPlas, Arnaud Joly, Brian
  Holt, and Ga{\"{e}}l Varoquaux.
\newblock {API} design for machine learning software: experiences from the
  scikit-learn project.
\newblock In {\em ECML PKDD Workshop: Languages for Data Mining and Machine
  Learning}, pages 108--122, 2013.

\bibitem[BORZ17]{Bailis2017InfrastructureFU}
Peter~D. Bailis, Kunle Olukotun, Christopher R{\'e}, and Matei~A. Zaharia.
\newblock Infrastructure for usable machine learning: The stanford dawn
  project.
\newblock {\em ArXiv}, abs/1705.07538, 2017.

\bibitem[BTY21]{bianchi-etal-2021-query2prod2vec}
Federico Bianchi, Jacopo Tagliabue, and Bingqing Yu.
\newblock {Q}uery2{P}rod2{V}ec: Grounded word embeddings for e{C}ommerce.
\newblock In {\em Proceedings of the 2021 Conference of the North American
  Chapter of the Association for Computational Linguistics: Human Language
  Technologies: Industry Papers}, pages 154--162, Online, June 2021.
  Association for Computational Linguistics.

\bibitem[BYT20]{bianchi2020bert}
Federico Bianchi, Bingqing Yu, and Jacopo Tagliabue.
\newblock Bert goes shopping: Comparing distributional models for product
  representations.
\newblock {\em arXiv preprint arXiv:2012.09807}, 2020.

\bibitem[BYT21]{tagliabue-etal-2021-bert}
Federico Bianchi, Bingqing Yu, and Jacopo Tagliabue.
\newblock {BERT} goes shopping: Comparing distributional models for product
  representations.
\newblock In {\em Proceedings of the 4th Workshop on e-Commerce and NLP}, pages
  1--12, Online, August 2021. Association for Computational Linguistics.

\bibitem[CAB{\etalchar{+}}22]{Naturearticle}
Patrick Chia, Giuseppe Attanasio, Federico Bianchi, Silvia Terragni, Ana
  Magalhães, Diogo Goncalves, Ciro Greco, and Jacopo Tagliabue.
\newblock Contrastive language and vision learning of general fashion concepts.
\newblock {\em Scientific Reports}, 12, 11 2022.

\bibitem[CAS16]{45530}
Paul Covington, Jay Adams, and Emre Sargin.
\newblock Deep neural networks for youtube recommendations.
\newblock In {\em Proceedings of the 10th ACM Conference on Recommender
  Systems}, New York, NY, USA, 2016.

\bibitem[CLB{\etalchar{+}}19]{10.1145/3292500.3330723}
Mia~Xu Chen, Benjamin~N. Lee, Gagan Bansal, Yuan Cao, Shuyuan Zhang, Justin Lu,
  Jackie Tsay, Yinan Wang, Andrew~M. Dai, Zhifeng Chen, Timothy Sohn, and
  Yonghui Wu.
\newblock Gmail smart compose: Real-time assisted writing.
\newblock In {\em Proceedings of the 25th ACM SIGKDD International Conference
  on Knowledge Discovery \&amp; Data Mining}, KDD '19, page 2287–2295, New
  York, NY, USA, 2019. Association for Computing Machinery.

\bibitem[CTB{\etalchar{+}}22a]{chia-etal-2022-come}
Patrick~John Chia, Jacopo Tagliabue, Federico Bianchi, Ciro Greco, and Diogo
  Goncalves.
\newblock {``}does it come in black?{''} {CLIP}-like models are zero-shot
  recommenders.
\newblock In {\em Proceedings of the Fifth Workshop on e-Commerce and NLP
  (ECNLP 5)}, pages 191--198, Dublin, Ireland, May 2022. Association for
  Computational Linguistics.

\bibitem[CTB{\etalchar{+}}22b]{10.1145/3487553.3524215}
Patrick~John Chia, Jacopo Tagliabue, Federico Bianchi, Chloe He, and Brian Ko.
\newblock Beyond ndcg: Behavioral testing of recommender systems with reclist.
\newblock WWW '22 Companion, page 99–104, New York, NY, USA, 2022.
  Association for Computing Machinery.

\bibitem[CYT21]{Chia2021AreYS}
Patrick~John Chia, Bingqing Yu, and Jacopo Tagliabue.
\newblock "are you sure?": Preliminary insights from scaling product
  comparisons to multiple shops.
\newblock {\em ArXiv}, abs/2107.03256, 2021.

\bibitem[Dav21]{cnn}
Kelly Davis.
\newblock {Accelerating ML within CNN}.
\newblock
  \url{https://medium.com/cnn-digital/accelerating-ml-within-cnn-983f6b7bd2eb/},
  2021.
\newblock [Online; accessed 19-Feb-2023].

\bibitem[DBJ22]{hbr2022}
Thomas~H. Davenport, Randy Bean, and Shail Jain.
\newblock Why your company needs data-product managers, 2022.

\bibitem[DCM{\etalchar{+}}12]{NIPS2012_6aca9700}
Jeffrey Dean, Greg Corrado, Rajat Monga, Kai Chen, Matthieu Devin, Mark Mao,
  Marc\textquotesingle~aurelio Ranzato, Andrew Senior, Paul Tucker, Ke~Yang,
  Quoc Le, and Andrew Ng.
\newblock Large scale distributed deep networks.
\newblock In F.~Pereira, C.J. Burges, L.~Bottou, and K.Q. Weinberger, editors,
  {\em Advances in Neural Information Processing Systems}, volume~25. Curran
  Associates, Inc., 2012.

\bibitem[{Dim}22]{datasurvery}
{Dimensional Research}.
\newblock What data scientists tell us about ai model training today, 2022.

\bibitem[DP12]{hbr2012}
Thomas~H. Davenport and DJ~Patil.
\newblock Data scientist: The sexiest job of the 21st century, 2012.

\bibitem[Eri22]{ericMLOPSMess}
Mihail Eric.
\newblock Mlops is a mess but that's to be expected, 2022.

\bibitem[Gan21]{23andme}
Manoj Ganesan.
\newblock {Developing safe and reliable ML products at 23andMe}.
\newblock
  \url{https://medium.com/23andme-engineering/machine-learning-eeee69d40736},
  2021.
\newblock [Online; accessed 19-Feb-2023].

\bibitem[GCSE22]{gong_abe_2022_7094191}
Abe Gong, James Campbell, Superconductive, and Great Expectations.
\newblock Great expectations, September 2022.
\newblock {If you use this software, please cite it using these metadata.}

\bibitem[GO12]{ReGithub}
Karan Goel and Laurel Orr.
\newblock Data-centric ai.
\newblock \url{https://github.com/HazyResearch/data-centric-ai}, 2012.

\bibitem[Hec21]{algo2019}
Lawrence~E. Hecht.
\newblock Add it up: How long does a machine learning deployment take?, 2021.

\bibitem[JdFMP13]{doi:10.1089/big.2013.0037}
Enric Junqu\'{e}~de Fortuny, David Martens, and Foster Provost.
\newblock Predictive modeling with big data: Is bigger really better?
\newblock {\em Big Data}, 1(4):215--226, 2013.
\newblock PMID: 27447254.

\bibitem[JT22]{dataOpsblog}
Hugo Bowne-Anderson Jacopo~Tagliabue.
\newblock Dataops and mlops for reasonable organizations, 2022.
\newblock [Online; accessed 19-Feb-2023].

\bibitem[MDM19]{DBLP:journals/corr/abs-1909-07930}
Piero Molino, Yaroslav Dudin, and Sai~Sumanth Miryala.
\newblock Ludwig: a type-based declarative deep learning toolbox.
\newblock {\em CoRR}, abs/1909.07930, 2019.

\bibitem[MIM15]{McSherry2015ScalabilityBA}
Frank McSherry, Michael Isard, and Derek~Gordon Murray.
\newblock Scalability! but at what cost?
\newblock In {\em HotOS}, 2015.

\bibitem[MR21]{10.1145/3475965.3479315}
Piero Molino and Christopher R\'{e}.
\newblock Declarative machine learning systems: The future of machine learning
  will depend on it being in the hands of the rest of us.
\newblock {\em Queue}, 19(3):46–76, jun 2021.

\bibitem[MSN{\etalchar{+}}19]{Mohan2019}
Vijai Mohan, Yiwei Song, Priyanka Nigam, Choon~Hui Teo, Weitian Ding, Vihan
  Lakshman, Ankit Shingavi, Hao Gu, and Bing Yin.
\newblock Semantic product search.
\newblock In {\em KDD 2019}, 2019.

\bibitem[Mur21]{realtor}
Nicole Murphy.
\newblock {Improving Data Science Processes to Speed Innovation at
  Realtor.com}.
\newblock
  \url{https://medium.com/realtor-com-innovation-blog/improving-data-science-processes-to-speed-innovation-at-realtor-com-b6b90fa530dc/},
  2021.
\newblock [Online; accessed 19-Feb-2023].

\bibitem[Ng21]{NG}
Andrew Ng.
\newblock Ai doesn’t have to be too complicated or expensive for your
  business, 2021.

\bibitem[Pet17]{gartnerenterprise}
Christy Pettey.
\newblock Artificial intelligence and the enterprise, 2017.

\bibitem[PGM{\etalchar{+}}19]{NEURIPS2019_9015}
Adam Paszke, Sam Gross, Francisco Massa, Adam Lerer, James Bradbury, Gregory
  Chanan, Trevor Killeen, Zeming Lin, Natalia Gimelshein, Luca Antiga, Alban
  Desmaison, Andreas Kopf, Edward Yang, Zachary DeVito, Martin Raison, Alykhan
  Tejani, Sasank Chilamkurthy, Benoit Steiner, Lu~Fang, Junjie Bai, and Soumith
  Chintala.
\newblock Pytorch: An imperative style, high-performance deep learning library.
\newblock In {\em Advances in Neural Information Processing Systems 32}, pages
  8024--8035. Curran Associates, Inc., 2019.

\bibitem[{Pre}22]{prefect}
{Prefect}.
\newblock Prefect, 2022.

\bibitem[Ren15]{43826}
David~K. Rensin.
\newblock {\em Kubernetes - Scheduling the Future at Cloud Scale}.
\newblock 1005 Gravenstein Highway North Sebastopol, CA 95472, 2015.

\bibitem[RM19]{10.1145/3299869.3320212}
Mark Raasveldt and Hannes M\"{u}hleisen.
\newblock Duckdb: An embeddable analytical database.
\newblock In {\em Proceedings of the 2019 International Conference on
  Management of Data}, SIGMOD '19, page 1981–1984, New York, NY, USA, 2019.
  Association for Computing Machinery.

\bibitem[SGHP22]{https://doi.org/10.48550/arxiv.2209.09125}
Shreya Shankar, Rolando Garcia, Joseph~M. Hellerstein, and Aditya~G.
  Parameswaran.
\newblock Operationalizing machine learning: An interview study, 2022.

\bibitem[SHG{\etalchar{+}}14]{43146}
D.~Sculley, Gary Holt, Daniel Golovin, Eugene Davydov, Todd Phillips, Dietmar
  Ebner, Vinay Chaudhary, and Michael Young.
\newblock Machine learning: The high interest credit card of technical debt.
\newblock In {\em SE4ML: Software Engineering for Machine Learning (NIPS 2014
  Workshop)}, 2014.

\bibitem[SKH{\etalchar{+}}21]{10.1145/3411764.3445518}
Nithya Sambasivan, Shivani Kapania, Hannah Highfill, Diana Akrong, Praveen
  Paritosh, and Lora~M Aroyo.
\newblock “everyone wants to do the model work, not the data work”: Data
  cascades in high-stakes ai.
\newblock In {\em Proceedings of the 2021 CHI Conference on Human Factors in
  Computing Systems}, CHI '21, New York, NY, USA, 2021. Association for
  Computing Machinery.

\bibitem[SKRC10]{5496972}
Konstantin Shvachko, Hairong Kuang, Sanjay Radia, and Robert Chansler.
\newblock The hadoop distributed file system.
\newblock In {\em 2010 IEEE 26th Symposium on Mass Storage Systems and
  Technologies (MSST)}, pages 1--10, 2010.

\bibitem[SMP19]{skelton2019team}
M.~Skelton, R.~Malan, and M.~Pais.
\newblock {\em Team Topologies: Organizing Business and Technology Teams for
  Fast Flow}.
\newblock G - Reference,Information and Interdisciplinary Subjects Series. IT
  Revolution, 2019.

\bibitem[{Spo}19]{luigi}
{Spotify Data Team}.
\newblock Luigi, 2019.

\bibitem[Sto12]{stonebraker12}
Michael Stonebraker.
\newblock What does 'big data' mean?, 2012.

\bibitem[Tag21]{10.1145/3460231.3474604}
Jacopo Tagliabue.
\newblock You do not need a bigger boat: Recommendations at reasonable scale in
  a (mostly) serverless and open stack.
\newblock In {\em Fifteenth ACM Conference on Recommender Systems}, RecSys '21,
  page 598–600, New York, NY, USA, 2021. Association for Computing Machinery.

\bibitem[Tag22a]{jacopo_medium}
Jacopo Tagliabue.
\newblock {Applied Research at Reasonable Scale}.
\newblock
  \url{https://medium.com/the-techlife/applied-research-at-reasonable-scale-8a74d2beed89},
  2022.
\newblock [Online; accessed 19-Feb-2023].

\bibitem[Tag22b]{towardDataScienceJT}
Jacopo Tagliabue.
\newblock Mlops without much ops, 2022.

\bibitem[TBS{\etalchar{+}}22]{https://doi.org/10.48550/arxiv.2207.05772}
Jacopo Tagliabue, Federico Bianchi, Tobias Schnabel, Giuseppe Attanasio, Ciro
  Greco, Gabriel de Souza~P. Moreira, and Patrick~John Chia.
\newblock Evalrs: a rounded evaluation of recommender systems, 2022.

\bibitem[TGR{\etalchar{+}}21]{CoveoSIGIR2021}
Jacopo Tagliabue, Ciro Greco, Jean-Francis Roy, Federico Bianchi, Giovanni
  Cassani, Bingqing Yu, and Patrick~John Chia.
\newblock Sigir 2021 e-commerce workshop data challenge.
\newblock In {\em SIGIR eCom 2021}, 2021.

\bibitem[TTGD21a]{DBLP:journals/corr/abs-2110-13601}
Jacopo Tagliabue, Ville Tuulos, Ciro Greco, and Valay Dave.
\newblock {DAG} card is the new model card.
\newblock {\em CoRR}, abs/2110.13601, 2021.

\bibitem[TTGD21b]{Tagliabue2021DAGCI}
Jacopo Tagliabue, Ville~H. Tuulos, Ciro Greco, and Valay Dave.
\newblock Dag card is the new model card.
\newblock {\em ArXiv}, abs/2110.13601, 2021.

\bibitem[Tuu21]{villeyoutube}
Ville Tuulos.
\newblock Metaflow: The ml infrastructure at netflix, 2021.

\bibitem[Tuu22]{VilleBook}
Ville Tuulo.
\newblock {\em Effective Data Science Infrastructure: How to make data
  scientists productive}.
\newblock Manning, 1st edition, 2022.

\bibitem[ZCD{\etalchar{+}}12]{10.5555/2228298.2228301}
Matei Zaharia, Mosharaf Chowdhury, Tathagata Das, Ankur Dave, Justin Ma, Murphy
  McCauley, Michael~J. Franklin, Scott Shenker, and Ion Stoica.
\newblock Resilient distributed datasets: A fault-tolerant abstraction for
  in-memory cluster computing.
\newblock In {\em Proceedings of the 9th USENIX Conference on Networked Systems
  Design and Implementation}, NSDI'12, page~2, USA, 2012. USENIX Association.

\end{thebibliography}

\end{document}